\documentclass[10pt,journal,compsoc]{IEEEtran}
\usepackage{booktabs}
\usepackage{color}
\usepackage{caption2}
\newcommand{\ygw}[1]{{\color{black} #1}}
\newcommand{\zwy}[1]{{\color{black} #1}}
\newcommand{\mtj}[1]{{\color{black} #1}}

\ifCLASSOPTIONcompsoc
  \usepackage[nocompress]{cite}
\else
  \usepackage{cite}
\fi
\ifCLASSINFOpdf
   \usepackage[pdftex]{graphicx}
   \graphicspath{{../pdf/}{../jpeg/}}
   \DeclareGraphicsExtensions{.pdf,.jpeg,.png}
\else
   \usepackage[dvips]{graphicx}
   \graphicspath{{../eps/}}
   \DeclareGraphicsExtensions{.eps}
\fi
\begin{document}
%
\title{Recursive-NeRF: An Efficient and Dynamically Growing NeRF}
%
%
%
%

\author{Guo-Wei~Yang,
        Wen-Yang~Zhou,
        Hao-Yang~Peng,
        Dun~Liang,
        Tai-Jiang~Mu,
        Shi-Min~Hu,~\IEEEmembership{Senior Member,~IEEE,}
\IEEEcompsocitemizethanks{
\IEEEcompsocthanksitem G.-W. Yang, W.-Y. Zhou, H.-Y. Peng, D. Liang, T.-J. Mu and S.-M. Hu are with the BNRist, Tsinghua University, Beijing 100084, China. 
\protect\\ Shi-Min Hu is the corresponding author. E-mail: shimin@tsinghua.edu.cn}
\thanks{Manuscript received May 17, 2021.}}

%
%

\markboth{Journal of \LaTeX\ Class Files,~Vol.~14, No.~8, August~2015}%
{Shell \MakeLowercase{\textit{et al.}}: Bare Demo of IEEEtran.cls for Computer Society Journals}
%



\IEEEtitleabstractindextext{%
\begin{abstract}
View synthesis methods using implicit continuous shape representations learned from a set of images, such as the Neural Radiance Field (NeRF) method, have gained increasing attention due to their high quality imagery and scalability to high resolution.
However, the heavy computation required by its volumetric approach prevents NeRF from being useful in practice; minutes are taken to render a single image of a few megapixels.
Now, an image of a scene can be rendered in a level-of-detail manner, so  we posit that a complicated region of the scene should be represented by a large neural network while a small neural network is capable of encoding a  simple region, enabling a balance between efficiency and quality. 
\emph{Recursive{-}NeRF} is our embodiment of this idea, providing an efficient and adaptive rendering and training approach for NeRF.
The core of Recursive{-}NeRF  learns uncertainties for query coordinates, representing the quality of the predicted color and volumetric intensity at each level.
Only query coordinates with high uncertainties are forwarded to the next level to a bigger neural network with a more powerful representational  capability.
The final rendered image is a composition of results from neural networks of all levels.
Our  evaluation  on three public datasets shows that Recursive{-}NeRF is \ygw{more efficient} than NeRF while providing state-of-the-art quality. \zwy{The code will be available at https://github.com/Gword/Recursive{-}NeRF.}
\end{abstract}

}

\maketitle

\begin{figure*}[!t]
  \centering\includegraphics[width=0.9\textwidth]{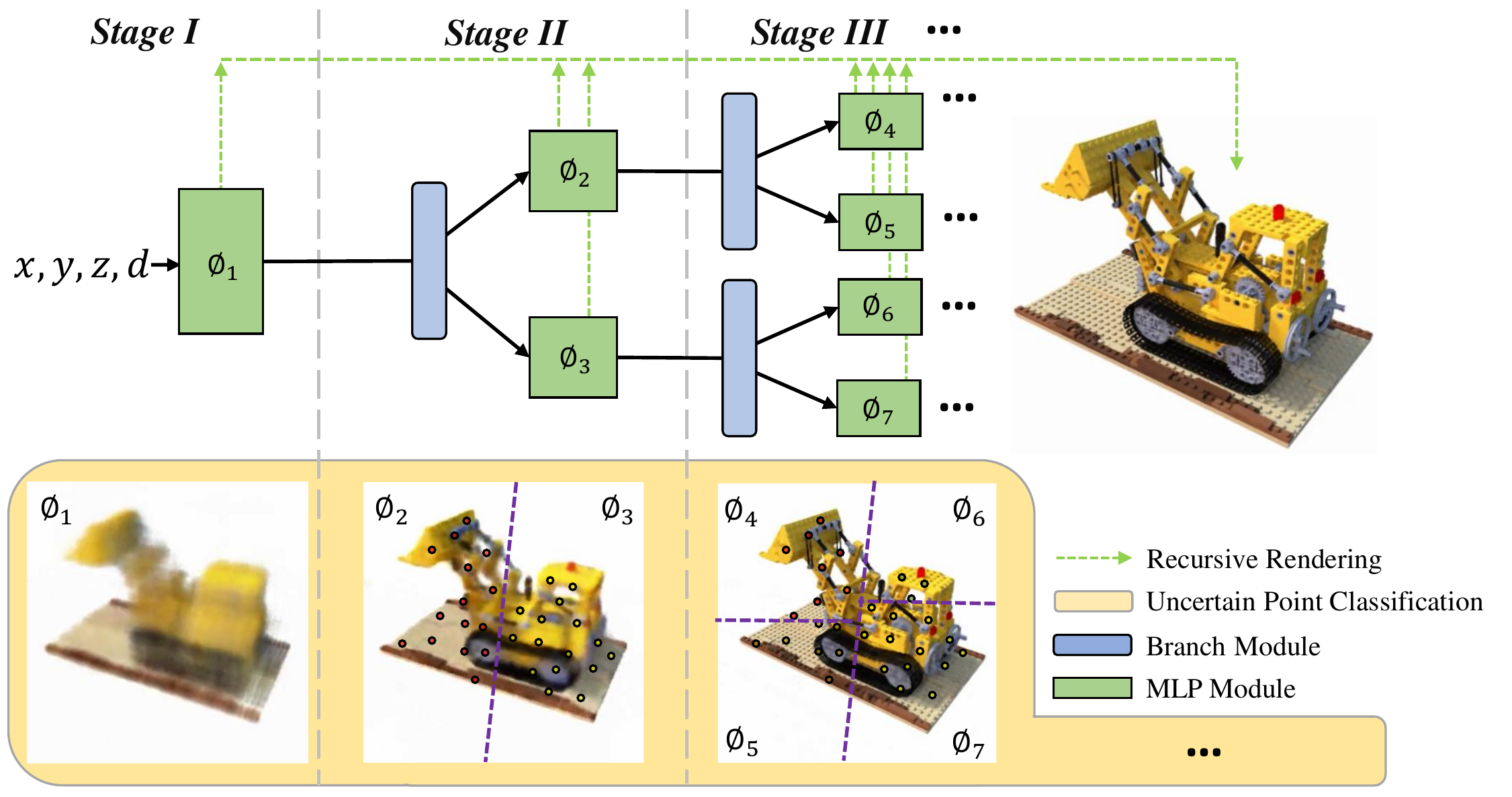}
  \caption{Pipeline of Recursive-NeRF. Given a position $(x,y,z)$ and viewing direction $d$, the initial network ${\Phi_1}$  outputs color $c_1$, density $\sigma_1$, and uncertainty $\delta_1$. All uncertain points are divided into several categories, and then ${\Phi_1}$ dynamically grows several branch networks to continue training for each subset of uncertain points until the network is believes that all points are reliably predicted. When rendering, early termination allows different points to exit at different times, reducing the  network load.}
  \label{fig:teaser}
\end{figure*}

\IEEEdisplaynontitleabstractindextext

%
\IEEEpeerreviewmaketitle

\IEEEraisesectionheading{\section{Introduction}\label{sec:introduction}}

Image-based rendering (IBR) is a popular topic in computer graphics with demonstrated value in virtual reality and deep learning for novel view synthesis and data augmentation.
The basic idea is to reconstruct the underlying geometry and appearance from a set of images, using representations which may be mesh-based~\cite{DBLP:journals/tog/Nimier-DavidVZJ19,DBLP:conf/iccv/Liu0LL19,DBLP:journals/tog/LiADL18,DBLP:conf/siggraph/BuehlerBMGC01}, volumetric~\cite{DBLP:journals/ijcv/SeitzD99,lombardi2019neural,DBLP:conf/cvpr/SitzmannTHNWZ19,mildenhall2019local,DBLP:journals/tog/PennerZ17,DBLP:journals/tog/ZhouTFFS18} or implicit~\cite{sitzmann2019scene,DBLP:conf/cvpr/NiemeyerMOG20}. These  are  used to synthesize novel views by interpolation~\cite{DBLP:conf/nips/ChenLGSLJF19} or rendering techniques~\cite{sitzmann2019scene,DBLP:conf/cvpr/NiemeyerMOG20}.
A recent development, the Neural Radiance Field (NeRF) method \cite{mildenhall2020nerf} implicitly encodes a scene or object using a fully-connected neural network, optimized by a naturally differentiable method. It provides excellent novel high-resolution photorealistic views  using a \emph{continuous} volumetric representation. It thus has been extended to large-scale scenes~\cite{zhang2020nerf++}, non-rigidly deforming scenes\ygw{~\cite{park2020deformable}}, dynamic lighting and appearance\ygw{~\cite{srinivasan2020nerv}}, etc.

Though NeRF  achieves unprecedented synthesis quality, its rendering process is extremely slow and makes high memory demands, so is unattractive for practical use. 
The bottleneck is the calculation of each pixel value by integrating along a rendering ray, which is approximated by hierarchical volume sampling in a similar way to importance sampling.
For each ray, NeRF samples 192 coordinates, each forward passing through the whole neural network, and in total millions of rays are required to render a single moderate-resolution image (say $800\times 800$ pixels).  
Previous work~\cite{liu2020neural} has improved NeRF's rendering speed by sampling more carefully using a sparse set of voxels, and avoiding evaluations on empty voxels. 


Let us consider some drawbacks of NeRF's rendering process. 
Firstly, NeRF uses the same network for all sample points, so NeRF encodes the whole scene using a single neural network model. 
For more complicated scenes, it is necessary to ensure the neural network has sufficient representational power. This is done by using an increased  number of network parameters, additional hidden layers, or increased dimensions of the latent vectors.
As a result, for every individual query sample, both the training and inference time of the network increase with greater scene complexity.
Furthermore, NeRF cannot self-adapt to scenes of different complexity. 
Secondly, regardless of whether inidividual query samples are complicated or simple, NeRF treats them equally and passes them through the entire neural network, which is overkill for regions which are empty or have simple geometric structures and textures.
These limitations seriously affect large-scale scene rendering.

Related problems in other rendering methods have been solved by using a level of detail (LOD) approach~\cite{LUEBKE20033}.
We suggest that it can be carried across to give an adaptive and efficient neural rendering approach based on NeRF: the rendered value of a sample needs to be further processed if and only its rendered quality is not high enough at the  current level of the neural network. 
Moreover, a coarser level may be represented with a smaller neural network while more detailed levels are  represented with larger neural networks.
The rendering of a sample adaptively passes through the neural network according to that sample's complexity.

We embody this concept in a method we call Recursive-NeRF (see Fig.~\ref{fig:teaser}) \zwy{, which recursively applies the NeRF structure with various number of linear layers in each stage when needed.} Starting with a small neural network, at each level, in additional to the color and volumetric intensity, Recursive-NeRF also predicts an uncertainty, indicating the quality of the current results. 
Recursive-NeRF then directly outputs results for those query coordinates in  the current level with low uncertainty, instead of passing them forward through the rest of the network.
Query coordinates with high uncertainty are forwarded in clusters to the next level, represented as multiple neural networks with more powerful representational capability.
The training process  terminates when the uncertainties for all query coordinates are less than a user-specified threshold, or some maximal number of iterations is reached.
In this way, Recursive-NeRF splits the work adaptively to decouple different parts of the underlying scene according to its complexity, helping to avoid unnecessary increase in network parameters.
Experiments demonstrates that Recursive-NeRF achieves significant gains in speed while providing high quality view synthesis.

In summary, our work makes the following main contributions:
\begin{itemize}
	\item a recursive scene rendering method, where early termination  prevents further processing once output quality is good enough, achieving state-of-the-art novel view synthesis results with much reduced computation, and
    \item a novel multi-stage dynamic growth method, which divides uncertain queries in the shallow part of the network, and continues to refine them in differently grown deep networks, making the approach adaptive for scenes with areas of differing complexity.
\end{itemize}

\section{Related Work}

Our approach  uses a neural 3D shape representation and dynamic neural networks for image based synthesis. A full review of these ideas is outside the scope of this paper, and we refer interested readers to~\cite{DBLP:journals/spic/ZhangC04} for classical IBR and ~\cite{NRSurvey20,DBLP:journals/corr/abs-2006-12057} for neural rendering.
We consider  the most closely related works below.

\subsection{Neural 3D shape representations for view synthesis}\label{sec_2_1}

Recently, there has been work training an MLP network to continuously represent a 3D scene, mapping  3D coordinates to an implicit representation, e.g.\ the signed distance function (SDF)~\cite{DBLP:conf/cvpr/JiangSMHNF20,DBLP:conf/cvpr/ParkFSNL19} or an occupancy field~\cite{DBLP:conf/cvpr/MeschederONNG19,DBLP:conf/cvpr/GenovaCSSF20}. Such approaches  usually need to be supervised with ground-truth 3D geometry.
An an alternative, learning  from a set of images has benefits, since images are more readily available, and supervision can be implemented with neural rendering techniques~\cite{DBLP:conf/cvpr/KatoUH18,DBLP:conf/iccv/Liu0LL19}.
\ygw{Scene Representation Networks (SRN)} \cite{sitzmann2019scene} uses an MLP network to learn scene geometry and appearance, proposing a differentiable ray-marcher to train the network end-to-end in an unsupervised manner.
\ygw{Neural Volumes (NV)} \cite{lombardi2019neural} learns a dynamic irregular warp field during ray-marching.
\ygw{Local Light Field Fusion (LLFF)} \cite{mildenhall2019local} expands each input view into a local light field through a multiplane image (MPI), then mixes adjacent local light fields to render novel views.

Recently proposed, NeRF \cite{mildenhall2020nerf} uses a sparse set of input views to optimize an MLP network which inputs a query point and outputs color and density. NeRF trains the network and renders the scene by sampling points in space by ray marching; it can generate high resolution images of high quality.
This approach has been adapted to handle more complicated scenarios.
For example, Zhang et al.~\cite{zhang2020nerf++} solve the parameterization problem arising when applying NeRF to object capture in 360$^\circ$ large-scale scenes.
Srinivasan et al.~\cite{srinivasan2020nerv} enhances NeRF for view synthesis under any lighting conditions.
Schwarz et al.~\cite{schwarz2020graf} propose generative radiance fields for 3D-aware image synthesis.
Park et al.~\cite{park2020deformable} optimize an additional continuous volumetric deformation field for non-rigidly deforming scenes.

However, these NeRF-based methods need a large number of samples in the rendering line of sight, so are slow.
NSVF~\cite{liu2020neural} focuses on the sampling strategy and introduces a sparse voxel octree on the basis of NeRF; speed is improved by avoiding calculating integrals in empty voxels.
Lindell et al.~\cite{lindell2020autoint} propose automatic integration to estimate volume integrals along the viewing ray in closed-form to avoid sampling, but their method suffers from quality degradation due to the piecewise approximation.

Although these methods have significantly accelerated NeRF, there is still much room for further improvement. Our dynamic network  adapts to the complexity of the scene,  significantly reducing the amount of calculation.
It is complementary to NSVF~\cite{liu2020neural}, and could be combined  for further speed.

\ygw{Recently, a series of good speed improvements have been achieved by caching neural network results~\cite{garbin2021fastnerf, yu2021plenoctrees}, but these methods will bring additional memory consumption. In the future, we will consider researching ways to improve rendering speed without additional memory consumption.}

\subsection{Dynamic Neural networks}

Dynamic neural networks dynamically adjust the network architecture according to equipment resources.

Multi-scale dense networks \cite{huang2017multi} train multiple classifiers according to different resource demands and adaptively apply them during testing.
\cite{yu2018slimmable} proposes switchable batch normalization and slimmable neural networks, which can adjust width according to  device resources.
\cite{yu2019universally} extends this idea to execute at arbitrary width.
\cite{yu2019autoslim} extends slimmable neural networks to change numbers of channels for better accuracy with constrained resources.

These approaches adjust the network architecture according to equipment resources, whereas we adaptively adjust the network architecture according to the network training situation. Also, previous dynamic networks solve a classification task\ygw{. Since the output of the classification network is the probability of predicting each category, the probability can naturally be used as the confidence.} Ours is a regression task, which determines whether the network is exited by predicting a confidence value. This is harder to train than a classification task.

\section{Analysis of Neural Radiance Fields}

\subsection{Neural Radiance Fields}

NeRF inputs continuous 5D coordinates, composed of a 3D position and a 2D viewing direction, and estimates view-dependent radiance fields and volume density at the corresponding position. By producing radiance fields, NeRF can simulate highlights and reflections well. NeRF calculates the color C(r) of a pixel in an image by integrating the ray from the camera to the pixel:
\begin{equation}
\label{eqn:01}
C(\mathbf{r})=\int_{0}^{+\infty}T(t)\sigma(r(t))c(\mathbf{r}(t),\mathbf{d})\,dt
\end{equation}
where $r(t)=\mathbf{o}+t\mathbf{d}$ is the camera ray emitted from $\mathbf{o}$ in direction $\mathbf{d}$, and $T(t)$ represents the cumulative transparency from $0$ to $t$:
\begin{equation}
\label{eqn:02}
T(t)=\exp\left(-\int_0^t\sigma(\mathbf{r}(s))\,ds\right).
\end{equation}
$c$ and $\sigma$ are directional emitted color and volume density which are calculated via an MLP network $F_\theta$:
\begin{equation}
\label{eqn:03}
F_\theta: (\mathbf{x},\mathbf{d})\rightarrow (c,\sigma)
\end{equation}

\subsection{Parameters and Scene Complexity}\label{cbpc}

\begin{figure}[!t]
  \centering
  \includegraphics[width=\linewidth]{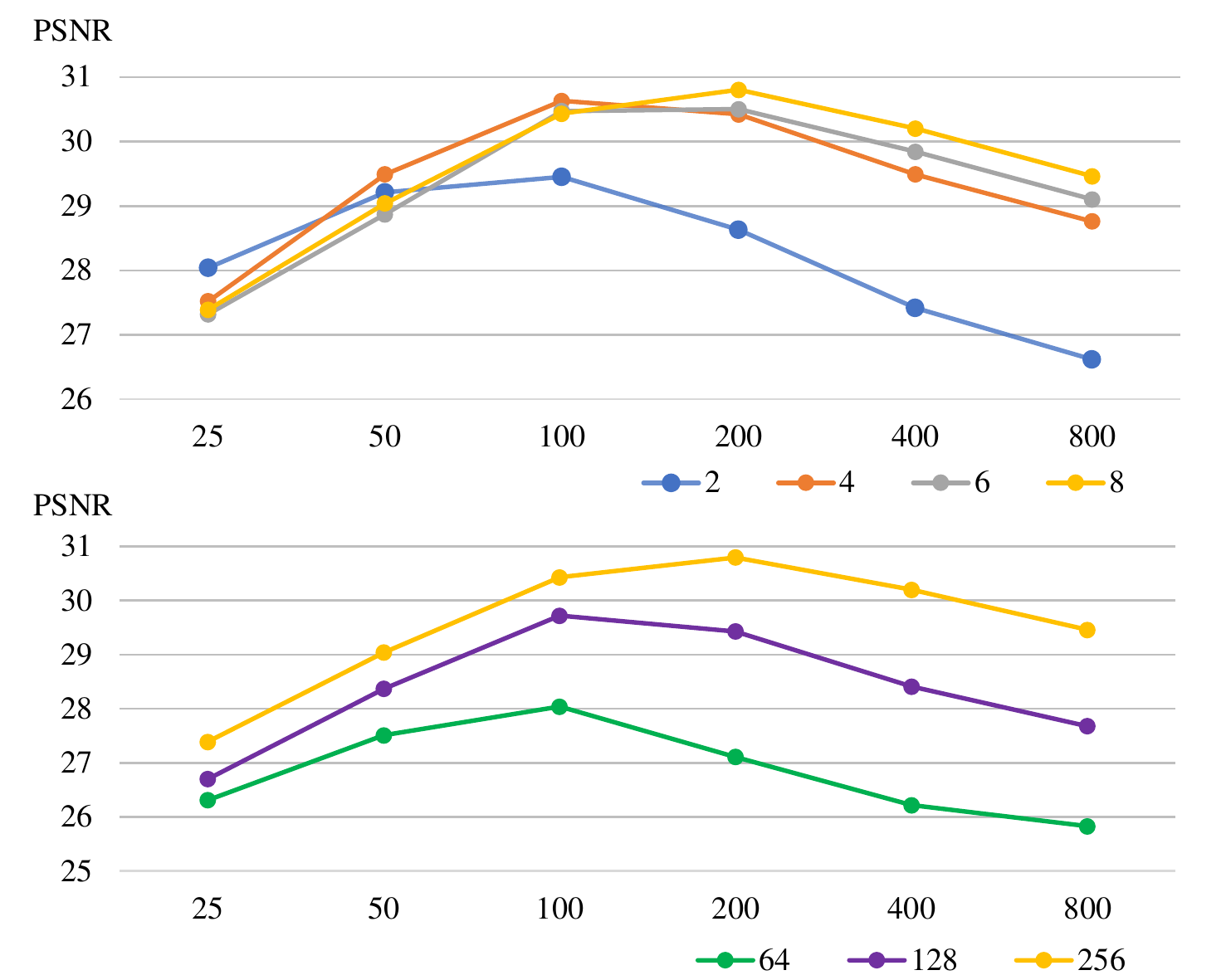}
  \caption{Correlation between parameters and scene complexity. \ygw{Different curves in the top and bottom plots represent various network depths and widths, respectively.} The horizontal axis is the resolution of the square image, representing the complexity of the model. \ygw{The vertical axis is the PSNR of the image, the higher the better.} }\label{fig:less}
\end{figure}


Scenes of greater complexity need to be represented using a larger number of parameters. At the same time, simple scenes can be represented by a small number of parameters.

We tested the PSNR of NeRF \cite{mildenhall2020nerf} on the Lego dataset for different numbers of network layers (2, 4, 6, 8), network widths (64, 128, 256) and image sizes (25, 50, 100, 200, 400, 800). The capacity of the network is positively corelated with the number of network layers and the network width. Here, image size is a proxy for complexity of the scene. \zwy{It can be seen that as the complexity of the scene increases, the PSNR first increases and then decreases. It is because when the scene is too simple, there is too little information for learning, while the scene will be 
beyond the capability of the network as it gets too complicated.}
It can be seen from Fig. \ref{fig:less} that when the scene is relatively simple, the representational capabilities of different networks are similar. As the scene becomes complex, the gap between the representational capabilities of different networks  widens. 

There is an intuitive solution: simply split the scene into several parts, with each part being represented by an identical individual network.
However, this solution has problems. Each part uses the same network architecture, while the complexity of different parts of the scene may differ, so ideally networks with different capabilities should be used to represent them. Furthermore, coarse-grained information  will be learned repeatedly.
 Recursive-NeRF overcomes both of these issues using a more sophisticated approach. 

\section{Recursive Neural Radiance Fields}

Recursive-NeRF use the NeRF approach in an LOD manner to adapt to the complexity of the underlying scene, which is trained in stages and changes dynamically, as shown in Fig. \ref{fig:teaser}.
At each stage, according to the predicted uncertainty, a query coordinate will be finalised or forwarded to the next stage which uses more powerful neural networks, controlled by an early termination mechanism.
All finalised predictions of color and intensity from each stage are gathered to render the final image.

In this section, we first introduce neural recursive fields (Sec. \ref{rnf}) which represent the whole scene from coarse to fine. 
Early termination (Sec. \ref{et}) allows our network to finalise the prediction when the uncertainty is low enough,  avoiding unnecessary calculations and speeding up rendering. 
We use the $k$-means algorithm to cluster the high uncertainty points in the current stage, thus dividing the scene into several parts for  finer-grained prediction. 
Additionally, the network grows several child branch networks to achieve dynamic growth (Sec. \ref{asd}). 
Overall, we recursively render (Sec. \ref{rr}) the entire scene, with input coordinates entering different branches for network prediction based on  previous clustering results. 

\subsection{Recursive Neural Fields}\label{rnf}

A recursive neural field takes its parent branch's output $y_{p_i}$ and the viewing direction $d$ as inputs, and predicts color $c_i$, density $\sigma_i$, uncertainty $\delta_{i}$ and a latent vector $y_i$:
\begin{equation}
\label{eqn:04}
F_{\Phi_i}: (y_{p_i},d)\rightarrow (c_i,\sigma_i,y_i,\delta_{i})
\end{equation}
where $F_{\Phi_i}$ represents the $i$th subnetwork. $F_{\Phi_1}$ is the root of our recursive network; in this case, $y_{p_i}$ is set to the query coordinate $(x,y,z)$.

\begin{figure}[!t]
  \centering
  \includegraphics[width=\linewidth]{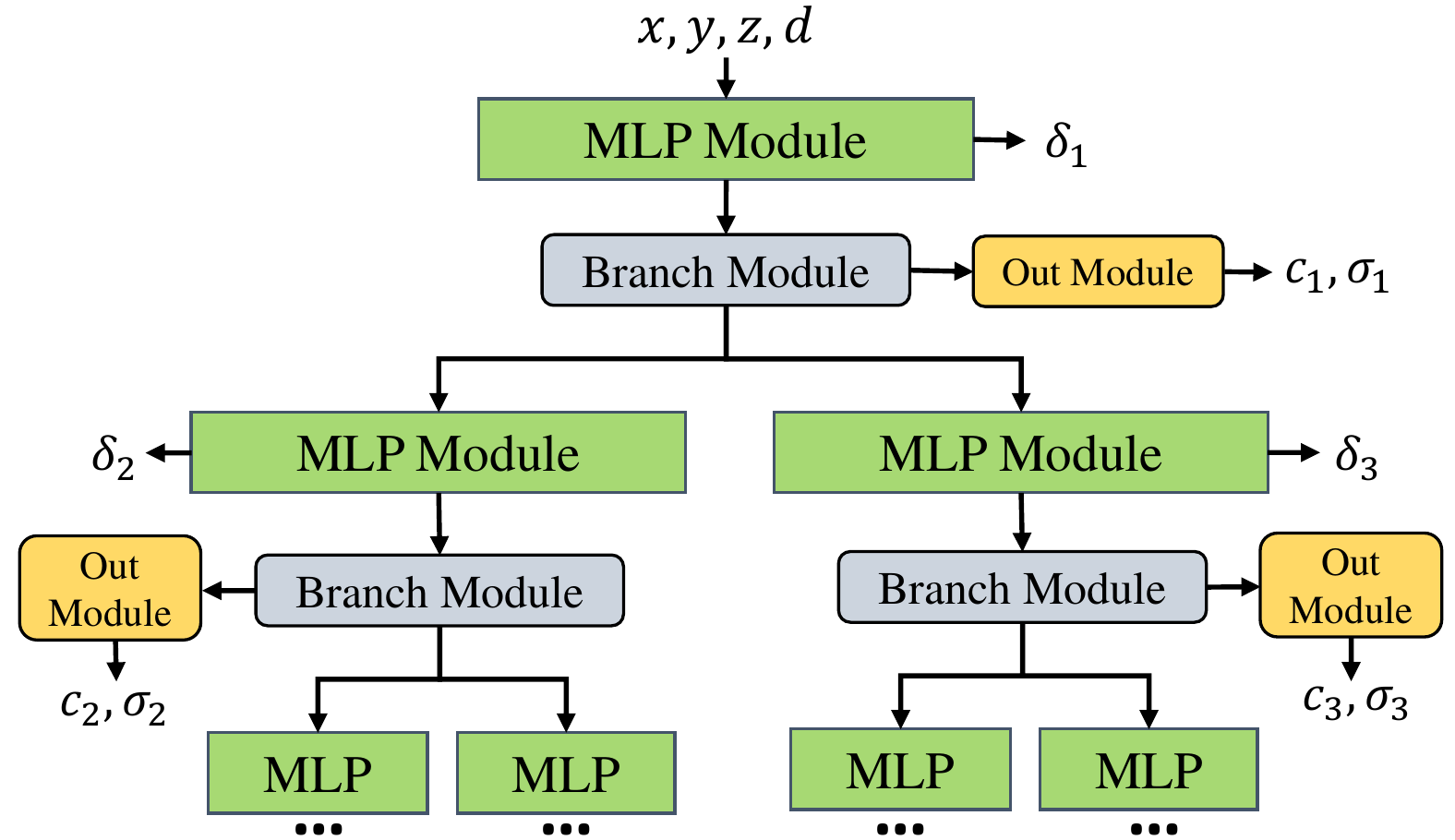}
  \caption{Network architecture of Recursive-NeRF. For every query $(x,y,z,d)$, the network predicts an uncertainty $\delta$ used to decide if the query should be finalised early. If so, it will enter the OutNet to predict its color $c$ and density $\sigma$. If not, the point split unit  determines which branch it should subsequently enter. }
  \label{fig:network}
\end{figure}

As shown in Fig.~\ref{fig:teaser} and~\ref{fig:network}, sub-network $F_{\Phi_i}$ consists of three main components: an MLP module, a Branch module and an Out module. The MLP module includes two or more Linear layers to ensure that the MLP module performs sufficiently complex processing of features. The Branch module predicts the uncertainty $\delta_{i}$ of each query point, forwards the points with low uncertainty to the Out module for output, and distributes points with high uncertainty to different sub-networks according to their distances to the $k_i$ cluster centers of $F_{\Phi_i}$. The Out module is responsible for decoding features into $c_i$ and $\sigma_i$. 

\subsection{Early Termination}\label{et}
Our early termination mechanism  allows the query coordinate to be finalised early (so not processed further)  when its predicted uncertainty is less than a certain threshold. We next present our novel uncertainty prediction method  for ray marching, then explain the special training method for Recursive-NeRF.

\subsubsection{Uncertainty prediction}

Each branch network predicts an uncertainty for the query coordinate, which we use to determine where the branch network exits.
 We use the original NeRF loss to help us predict uncertainty. NeRF adopts mean square error (MSE) between rendered images and real images as the loss for training coarse and fine networks:
\begin{equation}
\label{eqn:nerflossc}
\mathcal L_{MSEc}=\sum_{r\in \mathcal R}\left\|\hat C_c(r)-C(r)\right\|^2_2
\end{equation}
\begin{equation}
\label{eqn:nerflossf}
\mathcal L_{MSEf}=\sum_{r\in \mathcal R}\left\|\hat C_f(r)-C(r)\right\|^2_2
\end{equation}
where $\mathcal R$ contains rays in a mini-batch, $\hat C_c(r)$ and $\hat C_f(r)$ are rendered colors from coarse and fine networks, respectively, and $C(r)$ is the ground truth. Our coarse and fine networks have the same network structure and training. \zwy{The coarse and fine architecture is dedicated to sampling points since our network is designed to avoid unnecessary calculation for each sample point. Indeed, the previous uncertainty response can be used as sampling guidance for NeRF's fine network.} For simplicity, we no longer distinguish  coarse and fine networks, and use $\hat C(r)$ to represent the color rendered by any network.

We introduce two \zwy{regularization} losses to train the uncertainty effectively. We use a Linear layer following the output feature $y_{p_i}$ of $F_{\Phi_i}$ to calculate the uncertainty $\delta_{i}$. 
We use the squared error of the pixel to supervise $\delta_{i}$, the intent being that if a pixel has large error,  $\delta_{i}$ of the sample points that generate \ygw{the uncertainty associated to the sample points should also be large}. Therefore, we punish $\delta_{i}$ less than the squared error:
\begin{equation}
\label{eqn:07}
\mathcal L_{SE}=\sum_{r\in \mathcal R}\sum_{i=1}^N \ygw{max(E(r)-\delta_{c_i},0)}
\end{equation}
\begin{equation}
\label{eqn:08}
E(r)=\left\|\hat C(r)-C(r)\right\|^2_2
\end{equation}
where $E(r)$ is the squared error of ray $r$, $c_i$ is the sample point of ray $r$, and $\delta_{c_i}$ is the predicted uncertainty for query coordinates $c_i$.

To prevent $\delta_{i}$ from blowing up, we introduce another \zwy{regularization} loss: for every query point, we encourage $\delta_{i}$ to be as close to zero as possible:
\begin{equation}
\label{eqn:lossl0}
\mathcal L_{0}=\sum_{r\in \mathcal R}\sum_{i=1}^N \ygw{max(\delta_{c_i},0)}
\end{equation}

A weighted sum of $\mathcal L_{SE}$ and $\mathcal L_{0}$ gives the overall uncertainty loss:
\begin{equation}
\label{eqn:uncertainloss}
\mathcal L_{unct}=\alpha_1\mathcal L_{SE}+\alpha_2\mathcal L_{0}
\end{equation}
where $\alpha_1$ and $\alpha_2$ are weights, here  set to $\alpha_1=1.0$ and $\alpha_2=0.01$.

We use \zwy{regularization} loss instead of directly using $L_1$ loss to train  $\delta_{i}$ because the difficulty of accurately predicting $E(r)$ is about the same as directly predicting the color of the query coordinates for the neural network. In our network structure, it is difficult for a shallow network to have accurate $E(r)$. Therefore, we use \zwy{regularization} loss with unbalanced values for $\alpha_1$ and $\alpha_2$, so that the network can use larger penalties for points with uncertainty lower than loss, while uncertainty higher than loss will be less punished. In this way, the network learns the uncertainty into the upper bound of the complex loss function, so that only a truly certain point can terminate early.

\begin{figure}[!t]
  \centering
  \includegraphics[width=\linewidth]{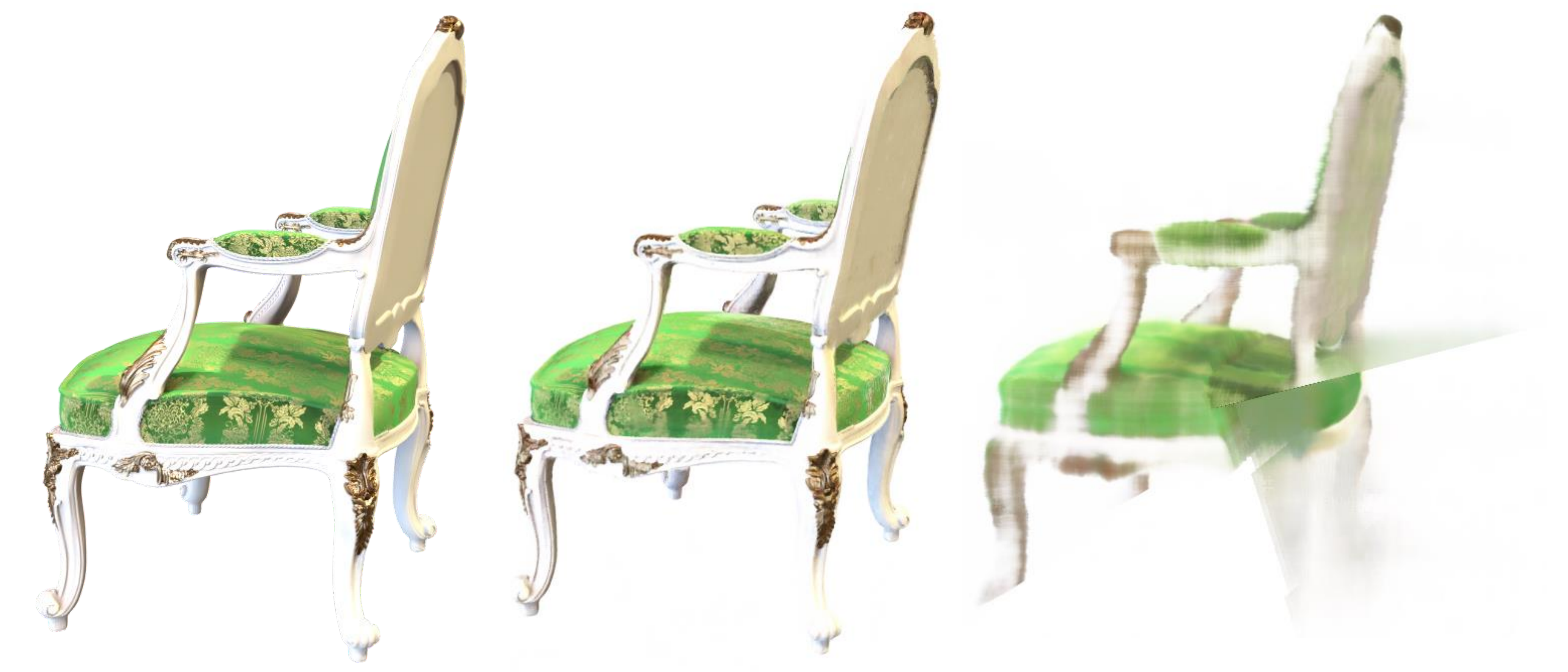}
  \caption{Alpha linear initialization comparison. Left: ground truth. Middle: our full model's result. Right: result of model without alpha linear initialization.}
  \label{fig:chair}
\end{figure}

\subsubsection{Multi-scale joint training}

We thus finalise queries with uncertainty lower than $\epsilon$ early, and forward points with uncertainty greater than or equal to $\epsilon$ to the deeper network. 
This early termination mechanism can reduce unnecessary calculations, but unfortunately,  also brings training difficulties. For a query coordinate $(x, y, z, d)$, the uncertainty may exceed $\epsilon$ at some stage, and the coordinate will be sent to a deeper network. However, if the deeper network has not been trained on this coordinate before, it will output an almost random value. This will cause great instability in the loss, affecting the training and may even  cause gradient explosion.

To solve this problem, we follow the practice in multi-scale dense networks~\cite{huang2017multi}: each time, all query coordinates are output through all outlets; images with early termination are also output, and their losses are weighted with equal weight during training.
Our overall loss function is thus:
\begin{equation}
\label{eqn:overallloss}
\mathcal L=\sum_{i=1}^D\beta_1\mathcal L_{MSE}^i+\beta_2 L_{unct}^i
\end{equation}
where $D$ is current number of stages (also $D-1$ times of network growth), 
and $L_{MSE}^i$ and $L_{unct}^i$ are the MSE loss and uncertainty loss of the output image of layer $i$, respectively. $\beta_1$ and $\beta_2$ are weights, set to $\beta_1=1.0$ and $\beta_2=0.1$.

\subsection{Dynamic Growth}\label{asd}
We now explain our adaptive dynamic growth strategy which clusters the uncertain queries at the current stage, and grows deeper networks according to the clustering result.

As shown in Fig.~\ref{fig:teaser}, in the initial stage, the network contains only one sub-network $\Phi_1$ which consists of two linear layers. After $I_1$ iterations of training for the initial network, we sample a number of points in space and calculate their uncertainties. We then cluster those points for which the uncertainty is higher than $\epsilon$; the clustering result determines the growth of the next stage network. To ensure that clustering is simple and controllable,  we use the $k$-means algorithm\zwy{, which can be replaced by a more efficient clustering algorithm} \mtj{such as methods in~\cite{xu2015comprehensive}}, with $k\in[2,4]$. The network  grows $k$ branches according to the cluster centers; these are e.g.\ $\Phi_2$ and $\Phi_3$. Downstream, query points are assigned to the branch with the closest cluster center. 

\zwy{When scenes become complicated, NeRF has to deepen its network, while we can simply add further branches to get the same result. There are two reasons why we split the grown network into two. Firstly, splitting the points will reduce the complexity of the network, otherwise, a deeper network is required for all points. Secondly, each child-network is only responsible for part of the scene independently, making it more effective and adaptive. Ablation experiment named ``no branching'' in Table \ref{tab:ablation} shows that our full model are better than using a single branch .}

\begin{figure}[!t]
  \centering
  \includegraphics[width=0.4\textwidth]{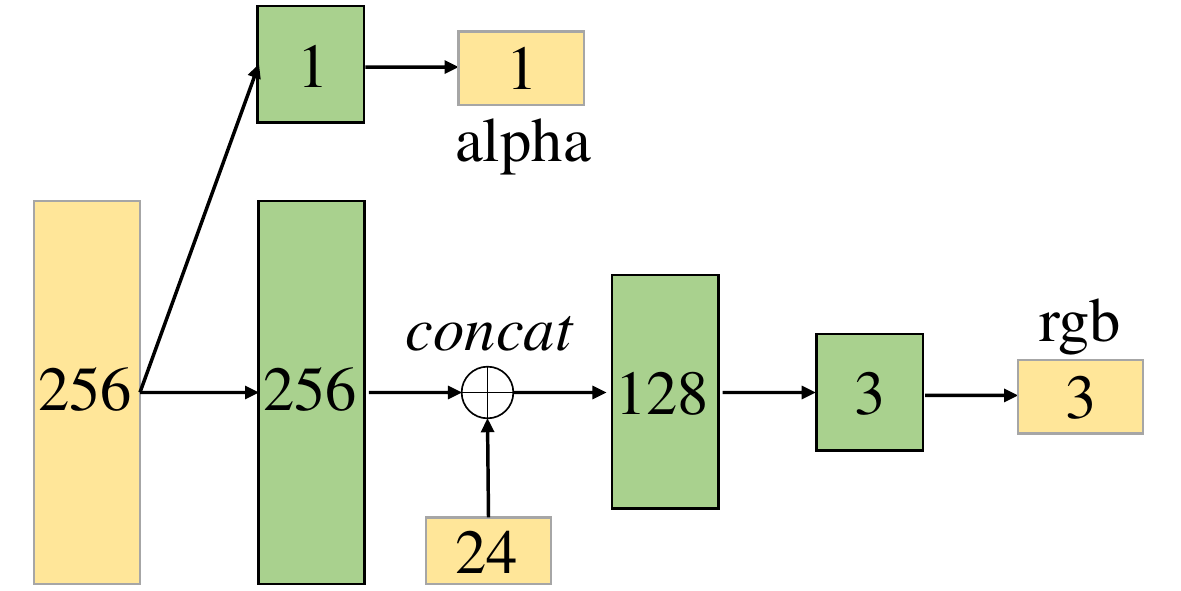}
  \caption{Left: network structure of OutNet. Green block: fully connected layer. Yellow blocks: input and output variables of the network.}
  \label{fig:outnet}
\end{figure}

\begin{figure*}[!t]
  \centering
  \includegraphics[width=1.0\textwidth]{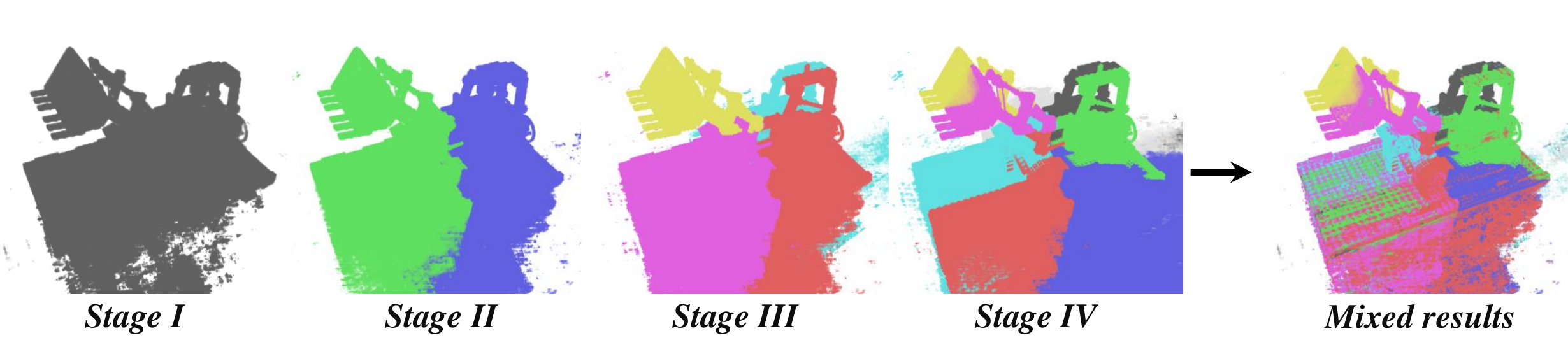}
  \caption{Scene segmentation at different stages. Left: segmentation at different stages. Right: blocks owning each query point, indicating early termination.}\label{fig:stage}
\end{figure*}

The growth-based network is trained for several iterations, and then clustered and grown. This process can be repeated successively until the uncertainty of most points is less than $\epsilon$. In order to finish  training in a reasonable amount of time, we specify that Recursive NeRF grows 
$3$ times in total. The value of $k$ used for each growth step can be different, but by default, we set $k$ for each to be $2$. 


\zwy{During training, sample points can exit at multiple stages, while points will exit only once at a specific stage during inferencing. Points found earlier to be reliable will immediately exit and be rendered. Which branch is taken depends on the results of clustering. We cluster the uncertain points in the current stage and feed them to different child-branches with the same structure in the next stage.}

Trials show that  direct growth of a randomly initialized network results in instability in the staged training, causing the density of some of the grown networks to reduce to 0. As a result the rendered scene can lack pieces,
 as shown in Fig. \ref{fig:chair}. The specific structure of the network's Outnet module is shown in Fig. \ref{fig:outnet}, where alpha linear is responsible for decoding features into the density of query points. \ygw{Our approach to overcoming this problem is to initialize the alpha linear weights of grown sub-network to be same as those of the parent.} This enables the density generation network of the subnet to inherit part of the density information of the parent, avoiding this instability.

We show staged clustering results for the Lego model in Fig. \ref{fig:stage}. It can be seen that the image includes finer and finer detail from the initial  to  final stage.



\begin{figure}[!t]
  \centering
  \includegraphics[width=\linewidth]{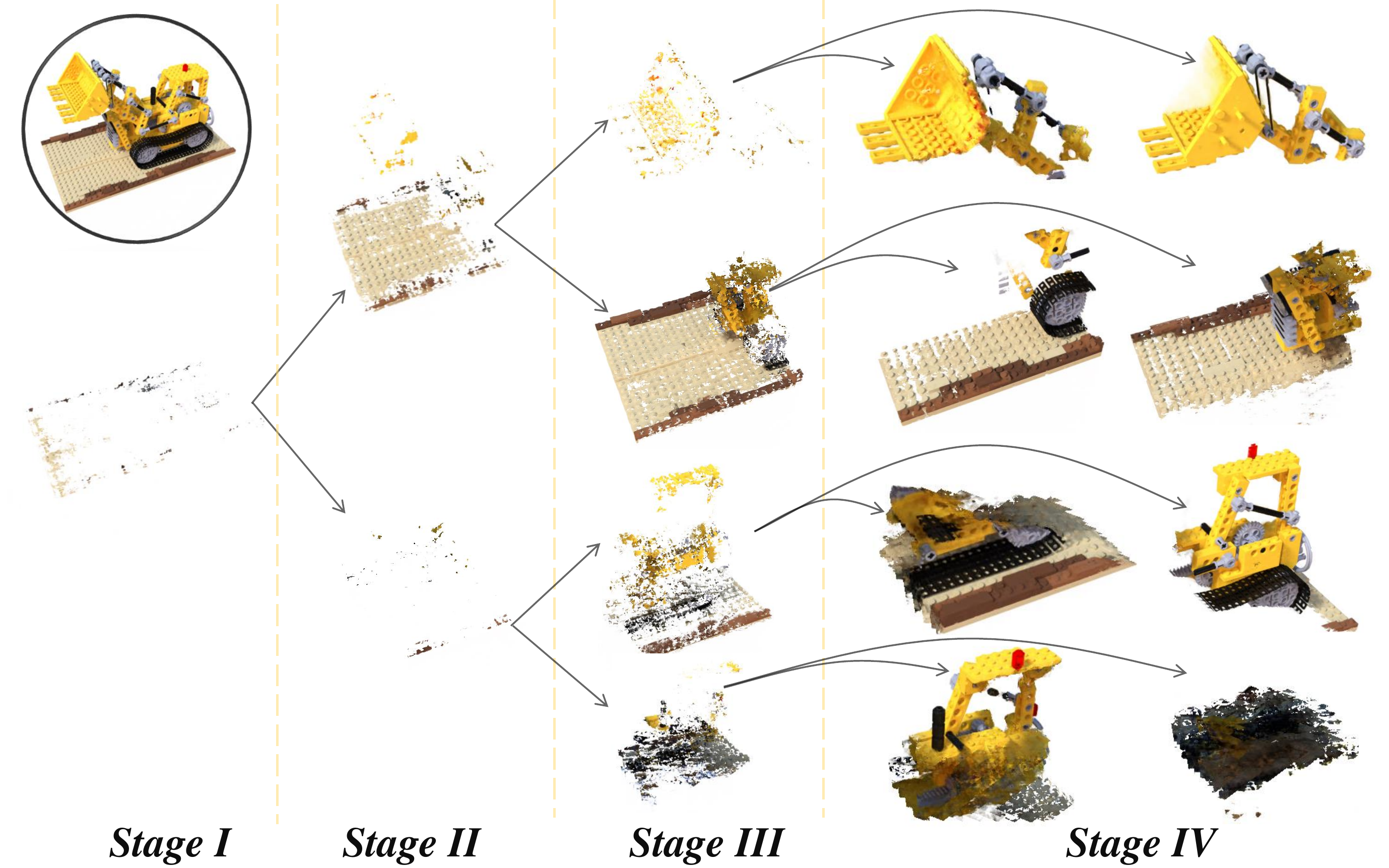}
  \caption{Recursive rendering. Various query points are finalised early in different stages, and finally all points are aggregated to form the  rendered image at the upper left.}
  \label{fig:recursiverender}
\end{figure}

\subsection{Recursive Rendering}\label{rr}

Unlike NeRF which outputs the color and density for all points in the last layer of the network, Recursive-NeRF renders the final image recursively. In the current view, all points whose uncertainty is lower than the threshold at the current stage can render a relatively fuzzy image. All points with uncertainty higher than the threshold  enter the next stage network to be further refined and other points of low uncertainty can exit from this stage. 
These points together with ones from all previous stages can render a clear image. 
 Fig. \ref{fig:recursiverender}
renders images using the points finalised at different stages,  these  being merged to form the final image at top-left. \zwy{The uncertainty is implicitly visualized in Fig. 7, where areas of low uncertainty at earlier stages are mainly \mtj{empty spaces} and surfaces with simple structure, such as the floor of the Lego model.}

Each input query point $r(t)$ exits from the first branch network in which its uncertainty probability is less than $\epsilon$. We use the color $c_i$ and density $\sigma_i$ predicted by the branch network to represent $c(r(t))$ and $\sigma(r(t))$ needed by Eq.~\ref{eqn:01}. Then we use Eq.~\ref{eqn:01} to calculate the color of the query point.
\begin{equation}
\label{eqn:recrender}
\sigma(r(t))=\sigma_i,i=min\{i|\delta_i<\epsilon \land r\in R_i\}
\end{equation}
\begin{equation}
\label{eqn:06}
c(r(t))=c_i,i=min\{i|\delta_i<\epsilon \land r\in R_i\}
\end{equation}
where $R_i$ is the set of points contained in the i-th branch.

\section{Experiments and discussion}

In this section, we first evaluate our Recursive-NeRF on different datasets and compare it with state-of-art alternatives.
Then we conduct ablation studies to validate the design choices of our approach, including early termination, uncertain point clustering and the branching mechanism.


\subsection{Experimental Settings}

\subsubsection{Deep Learning Framework} 
All of our experiments were implemented using the Jittor deep learning framework~\cite{hu2020jittor}.  
Jittor supports dynamic graph execution, allowing the neural network to be dynamically changed during each training stage, so is well suited to training our Recursive-NeRF.

\subsubsection{Datasets}  We evaluated our method on Synthetic-NeRF~\cite{mildenhall2020nerf}, and Cornell Box dataset~\cite{cornellbox}. 
%
\textbf{Synthetic-NeRF} contains eight man-made objects with complicated geometry and materials. Each object is realistically rendered in 300 views {at a resolution of $800 \times 800$ pixels}. We used the same split into training and testing data as NeRF~\cite{mildenhall2020nerf}.
%
\textbf{Cornell Box} {is a relatively simple synthetic scene, {mainly composed of boxes}. We adopted this dataset to demonstrate the effectiveness of our method.} 
We rendered 400 pictures at a resolution of $800\times 800$ pixels {from views uniformly sampled with the camera moving along a spiral
curve}, and randomly selected 200 pictures as the training set, and the remainder for testing.


\subsubsection{Baseline} Using the above datasets, we compared our approach to several current state-of-the-art methods: \ygw{SRN} \cite{sitzmann2019scene}, \ygw{NV} \cite{lombardi2019neural}, \ygw{LLFF} \cite{mildenhall2019local}, and NeRF \cite{mildenhall2020nerf}. \zwy{As explained in Sec. \ref{sec_2_1}, NSVF focuses on avoiding calculating integrals for empty voxels by using a more reasonable sampling strategy, while each sampled point still goes through \mtj{a whole} NeRF. Recursive-NeRF improves upon NeRF in a complementary way by avoiding unnecessary calculation for each sample point by using an adaptive early termination strategy based on the learned uncertainty, which can be combined with NSVF to further improve the original NeRF. We thus do not make a  comparison between ours and NSVF in what follows.}
	
\subsubsection{Implementation details}

\begin{figure*}[!t]
  \centering
  \includegraphics[width=1\textwidth]{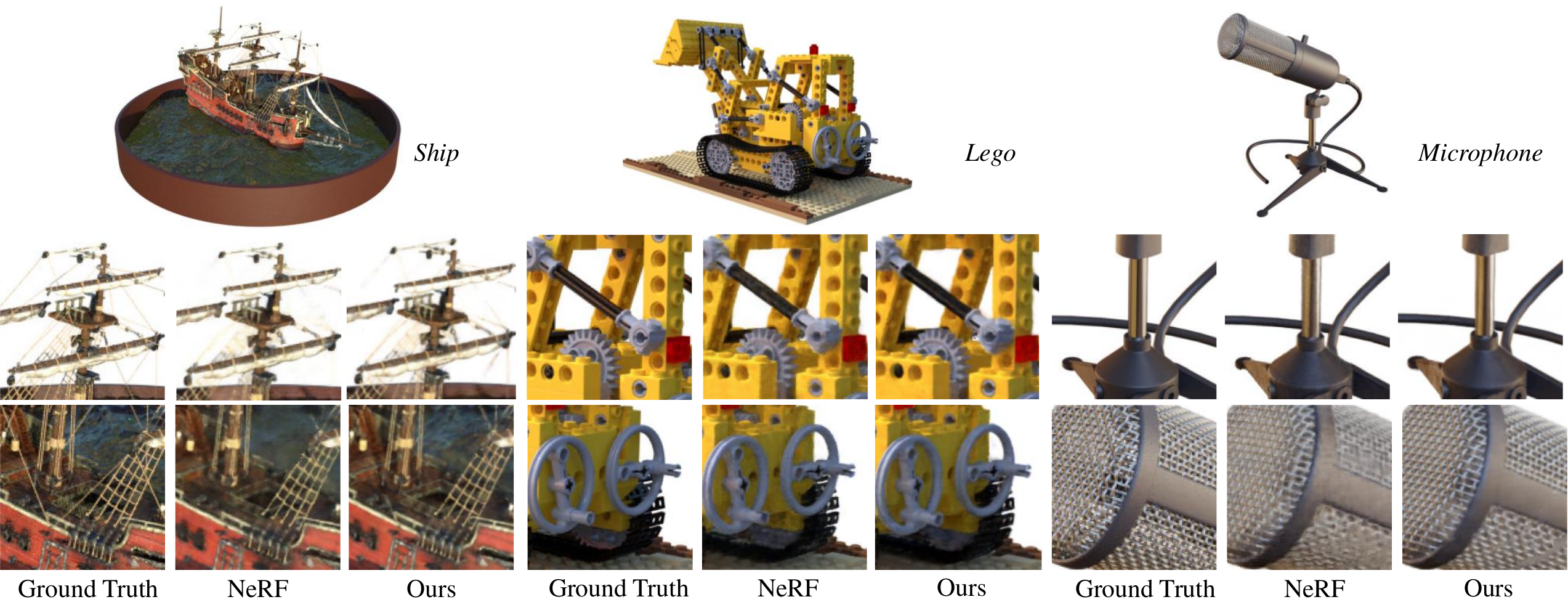}
  \caption{Qualitative results. Top: scene. Middle, below: Two close-ups of the scene. We show the ground truth, the results of NeRF rendering, and of our method in turn.}\label{fig:qualitative }
\end{figure*}

The batchsize {was set to 4096 in our training stage}, 64 and 128 sampling points were used for the coarse  and  networks respectively, and each model was trained for 300k iterations, all as in the NeRF paper. Our network underwent four stages of training, as it was grown three times. The initial and grown networks had 2, 2, 4, and 4 linear layers respectively. We used Adam as our optimizer and set a learning rate with initial value of $5\times 10^{-4}$ and 10 times exponential learning rate decay after 250k iterations.
%



\subsection{Results}

\begin{figure}[!t]
  \centering
  \includegraphics[width=\linewidth]{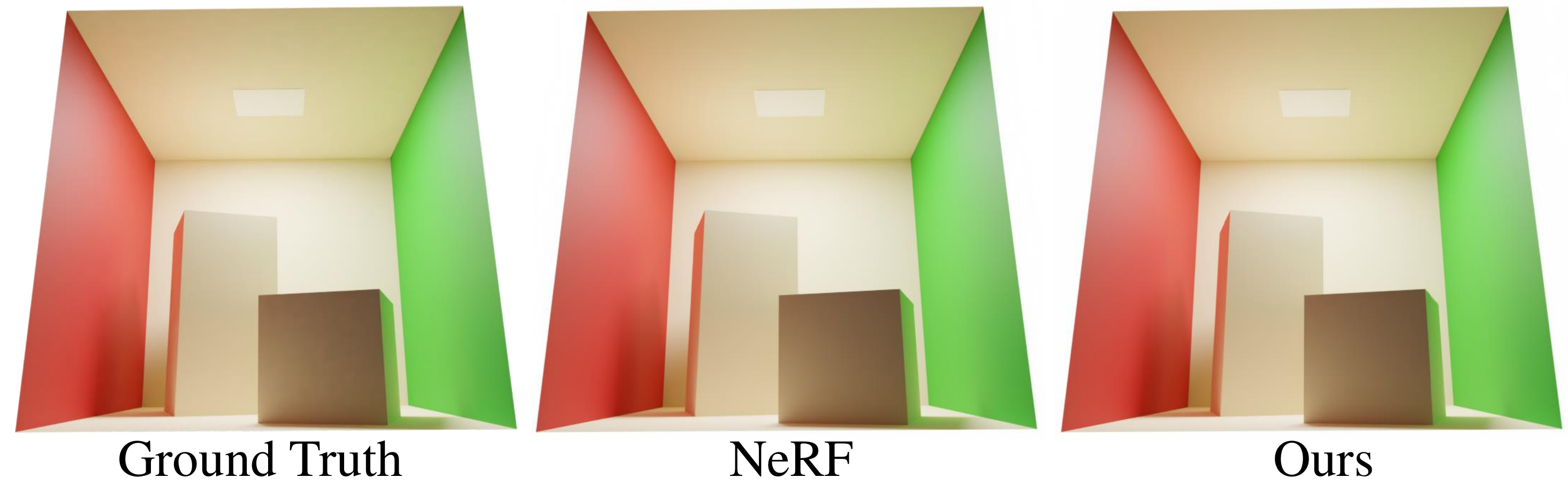}
  \caption{Qualitative comparison for Cornell Box. Left: original image . Middle: NeRF's result. Right: our result.}\label{fig:cornellboxresult}
\end{figure}


\begin{figure}[!t]
  \centering
  \includegraphics[width=\linewidth]{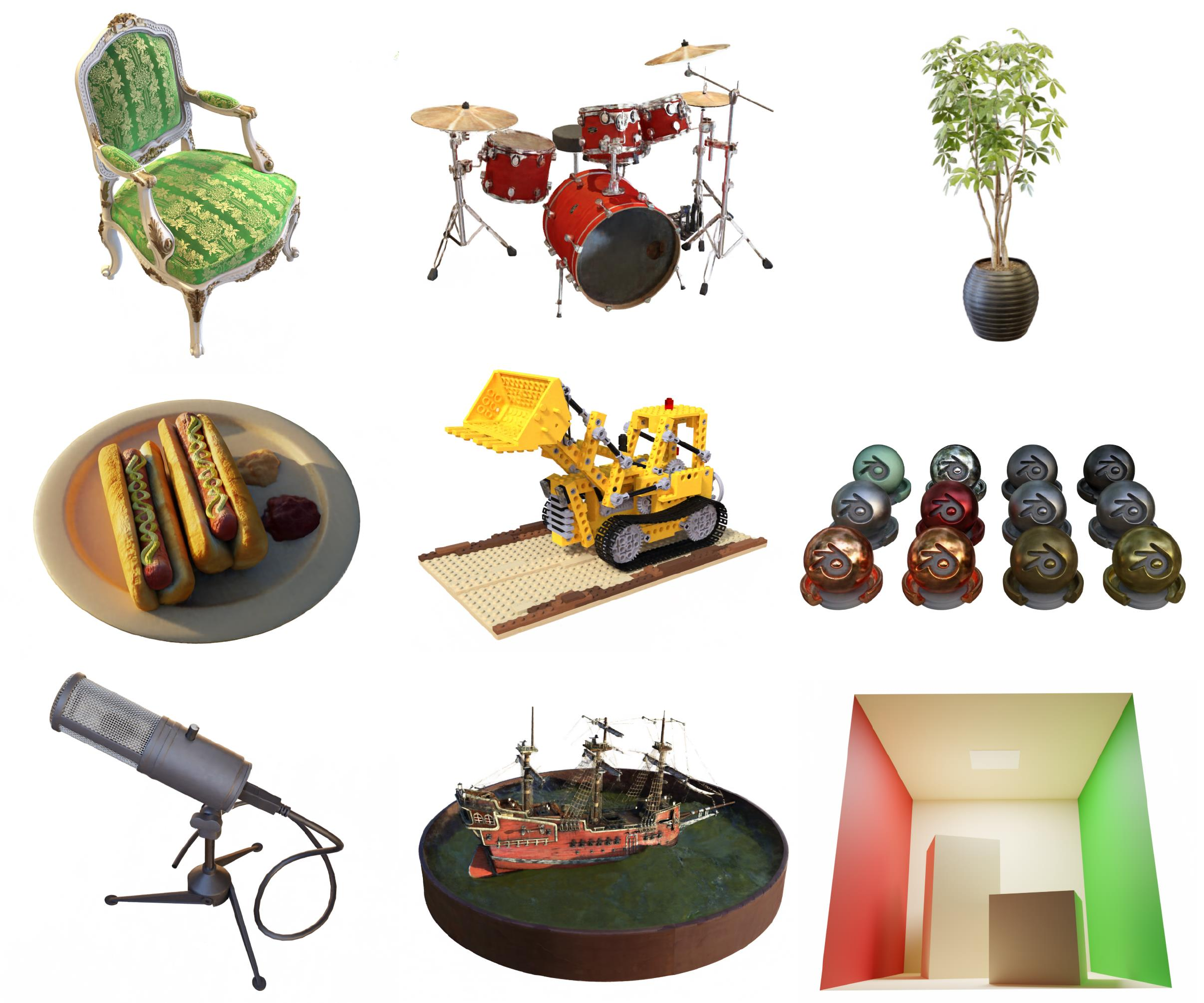}
  \caption{Further rendering results from Recursive-NeRF.}\label{fig:moreresults}
\end{figure}

\subsubsection{Qualitative comparison} We show different rendering results on Synthetic-NeRF dataset in Fig. \ref{fig:qualitative }. 
Comparative results on the Cornell Box dataset are shown in Fig. \ref{fig:cornellboxresult}. 
Our method generates {much clearer} local details {than the baseline NeRF} on the Synthetic-NeRF dataset and achieves comparable results on Cornell Box dataset.
Fig. \ref{fig:moreresults} galleries more results Recursive-NeRF renders at other viewpoints on both Synthetic-NeRF and Cornell Box dataset.

\begin{table}%
\caption{Quantitative Comparison on the Synthetic-NeRF dataset~\cite{mildenhall2020nerf}}
\label{tab:synthetic}
\begin{minipage}{\columnwidth}
\begin{center}
\begin{tabular}{lrrrrr}
  \toprule
  Method & PSNR$^\uparrow$ & SSIM$^\uparrow$ & LPIPS$^\downarrow$ & FLOPs$^\downarrow$ & Time$^\downarrow$\\ \midrule
  SRN \cite{sitzmann2019scene}      & 22.26 & 0.846 & 0.170 & - & -\\
  NV \cite{lombardi2019neural}  & 26.05 & 0.893 & 0.160 & - & -\\
  LLFF \cite{mildenhall2019local}     & 24.88 & 0.911 & 0.114 & - & -\\
  NeRF \cite{mildenhall2020nerf}    & 31.01 & 0.947 & 0.081 & 591488 & \ygw{26.14}\\
  \midrule
   Ours & \textbf{31.34} & \textbf{0.953} & \textbf{0.052} & \textbf{377159} & \ygw{\textbf{17.68}}\\
  \bottomrule
\end{tabular}
\end{center}
\bigskip\centering
\end{minipage}
\end{table}

\begin{table}%
\caption{Quantitative Comparison on Cornell Box}
\label{tab:cornell}
\begin{minipage}{\columnwidth}
\begin{center}
\begin{tabular}{llllll}
  \toprule
  Method & PSNR$^\uparrow$ & SSIM$^\uparrow$ & LPIPS$^\downarrow$ & FLOPs$^\downarrow$ & Time$^\downarrow$\\ \midrule
  NeRF & 49.237 & 0.996 & 0.015 & 591488 & \ygw{26.14}\\
  Ours & 48.010 & 0.996 & 0.010 & 198361 & \ygw{10.82}\\
  \bottomrule
\end{tabular}
\end{center}
\bigskip\centering
\end{minipage}
\end{table}

\begin{figure*}[!t]
  \centering
  \includegraphics[width=1\textwidth]{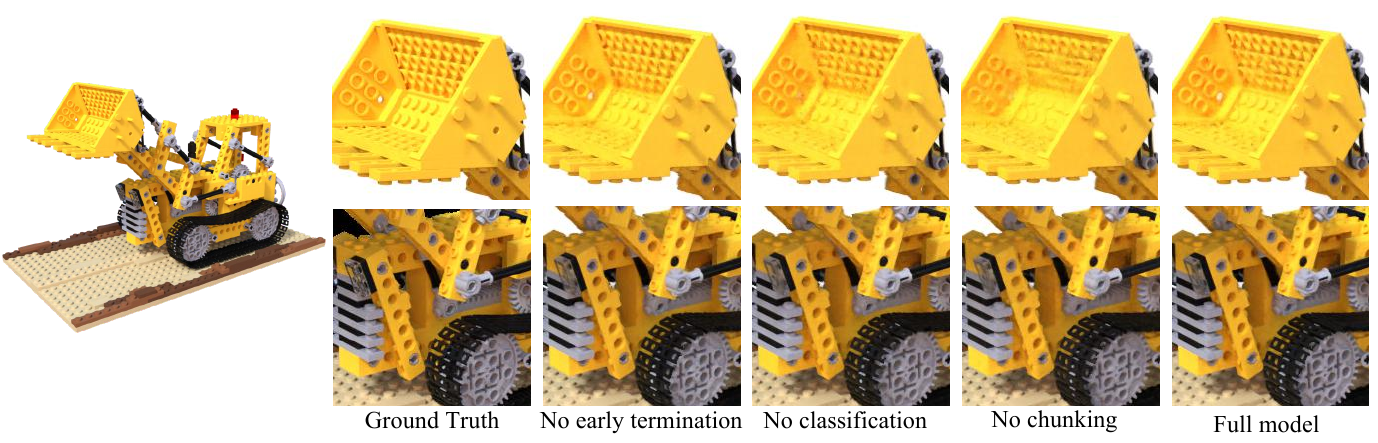}
  \caption{Qualitative comparison of ablation experiment.}\label{fig:ablation}
\end{figure*}
	
\subsubsection{Quantitative comparison} We also used PSNR and SSIM to  enable a quantitative comparison of the results (higher is better), as well as LPIPS~\cite{DBLP:conf/cvpr/ZhangIESW18} (lower is better). 
Results for the Synthetic-NeRF dataset are shown in Table \ref{tab:synthetic}, and demonstrate that our method can perform well on the general dataset. Results for the Cornell Box dataset are shown in Table \ref{tab:cornell}. We have reduced the amount of calculation by about 2/3, with only a slight loss of accuracy.



\subsubsection{Speed comparison} We show the number of floating point operations FLOPs \ygw{and rendering time} in both Table~\ref{tab:synthetic} and~\ref{tab:cornell}. Our network \ygw{requires fewer operations} for both simple and complex scenes, with greater speed improvement for the simple scenes as to be expected.
The maximum depth of our network can reach 12 layers, which is deeper than NeRF's 8 layers. Using on our early termination strategy, we are able to finalise a large number of simple query points in the shallow part of the network, leaving the deep network to focus on the complex information, thereby providing better results. Although the number of deepest layers in our network is greater,  our early termination strategy results in Recursive-NeRF reducing  NeRF's computational effort by $37\%$\ygw{, and $32.36\%$ less time to render a image at the same time}.

\zwy{\subsubsection{Distribution of sample termination} The distribution of sample point termination shows that the ratios of points terminated in the 2nd, 4th, 8th, and 12th layers are 45.3\%, 27.9\%, 7.2\%, and 19.6\%, respectively. The sample points will go through 4.95 layers on average in our network, while 8 layers are required in NeRF, and only 19.6\% of points go through a deeper network than in NeRF. Thus, our adaptive approach can effectively reduce the computation according to the learned uncertainty: it is more than a simply deeper NeRF.}


\subsection{Ablation Study}\label{abla}

We conducted ablation experiments as described below; the results are shown in Table. \ref{tab:ablation}. A qualitative comparison of results on the Lego dataset is shown in Fig. \ref{fig:ablation}.

\subsubsection{Effect of early termination} Early termination is a key part of our method. We trained our model without early termination\zwy{, which means all sampling points leave at the last exit}. It can be seen that the amount of computation increased significantly. The early termination mechanism greatly improves performance and only causes a very minor  degradation of the results.

\begin{table}%
\caption{Ablation experiment}
\label{tab:ablation}
\begin{minipage}{\columnwidth}
\begin{center}
\begin{tabular}{lrrrr}
  \toprule
  Method & PSNR$^\uparrow$ & SSIM$^\uparrow$ & LPIPS$^\downarrow$ & FLOPs$^\downarrow$\\ \midrule
  No classification      & 32.697 & 0.966 & 0.035 & 326463\\
  No branch & 32.374 & 0.962 & 0.040 & 349504\\
  No early termination & 33.118 & 0.970 & 0.032 & 854656\\
  \midrule
  Full model & 32.900 & 0.967 & 0.033 & 347765\\
  \bottomrule
\end{tabular}
\end{center}
\bigskip\centering
\end{minipage}
\end{table}

\begin{figure}[!t]
  \centering
  \includegraphics[width=\linewidth]{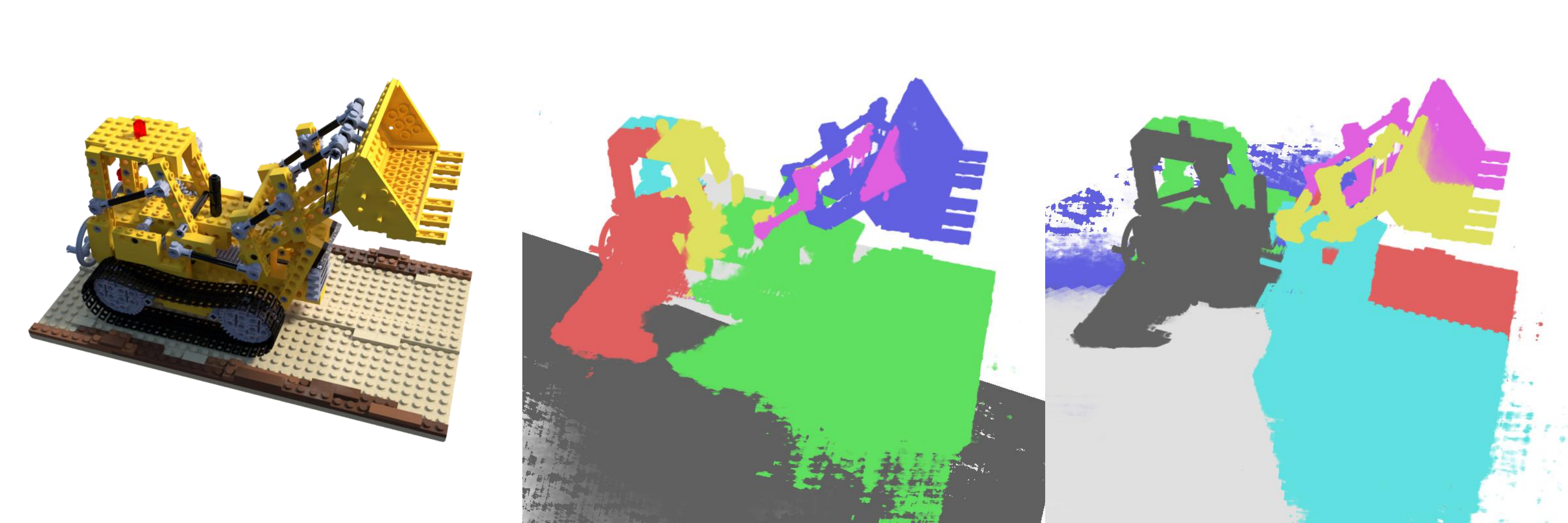}
  \caption{Scene segmentation results. Left: original  Lego scene. Middle: result using random division (no clustering). Right: result of clustering uncertain points using $K$-means.}\label{fig:kmeans}
\end{figure}

\subsubsection{Effect of uncertain point classification} We use $K$-means to divide the scene into blocks. As an alternative, we randomly divided the scene into blocks and compared the outcomes.  Fig. \ref{fig:kmeans}  shows the results. Without clustering, many blocks  contain many discontinuous parts, and the block size is also uneven, reducing the quality of the final image.

\subsubsection{Effect of branching} 
To demonstrate the effectiveness of our network block structure, we compare the results with a chain structure of the same depth network (No branch in Table \ref{tab:ablation}). The chain network had a $\#2FC-\#2FC-\#4FC-\#4FC$ structure where $\#iFC$ represents a fully connected layer with i-layers. The query points could terminate early from the 2nd, 4th, 8th, and 12th levels.
The branching strategy divides  complex parts into different branch networks for learning, allowing the network to decouple complex scenarios and conduct targeted learning. Thus branch strategy brings a significant increase in quality.



\subsection{Limitations and future works}

Recursive-NeRF rendering for large-scale scenes is still challenging. Inaccurate camera position and motion blur will degrade the performance of both Recursive-NeRF and NeRF, which restricts their application to real scenes. Although the speed of Recursive-NeRF  is better than that of the original  NeRF, it is still not adequate for real-time use. We hope to further improve the performance of Recursive-NeRF, to adapt it to more complex scenes.

{In principle Recursive-NeRF can add as many stages as desired to represent a large and complicated scene.
Currently, however, the maximal number of stages is fixed to 4. Also, the injection of sub-networks for a query coordinate is also fixed during training once the cluster centers of uncertain points at each stage has been determined, making it unadjustable in later training processes.
This could be improved in two ways to make our approach more suitable for large and complicated scenes. 
Firstly, we could select an appropriate number of stages to best represent the whole scene by using techniques from neural architecture search~\cite{DBLP:journals/jmlr/ElskenMH19}.
Secondly, learnable clustering~\cite{DBLP:conf/cvpr/WilliamsPNPYT20} of uncertain points could be exploited to make the branched networks more adaptive to complexity of parts of the underlying scene.}

\section{Conclusions}  

In this paper, we have proposed the idea of adaptively modeling parts of a scene with different complexity using neural networks of different representation capability, analogous to level of detail. We have  used it to construct a dynamically growing neural network for novel view synthesis, called Recursive-NeRF.
It extends  basic NeRF by additionally predicting  uncertainty for the results, and uses it to dynamically branch new, more powerful neural networks to represent  more uncertain regions, allowing it to efficiently learn  implicit geometric and appearance representations for complicated scenes.
Our experiments have demonstrated the effectiveness of our method and show that, compared to NeRF, Recursive-NeRF can generate more photorealistic views in a more efficient computation.

\ifCLASSOPTIONcompsoc
  \section*{Acknowledgments}
\else
  \section*{Acknowledgment}
\fi

\ygw{This work was supported by the Natural Science Foundation of China (Project Number 61521002). We would like to thank Guo-Ye Yang for his kindly help in experimentation and  Prof. Ralph R. Martin for his help in writing.}

\ifCLASSOPTIONcaptionsoff
  \newpage
\fi



%

%

\bibliographystyle{IEEEtran}
\bibliography{bibliography}



\begin{IEEEbiography}[{\includegraphics[width=1in,height=1.25in,clip,keepaspectratio]{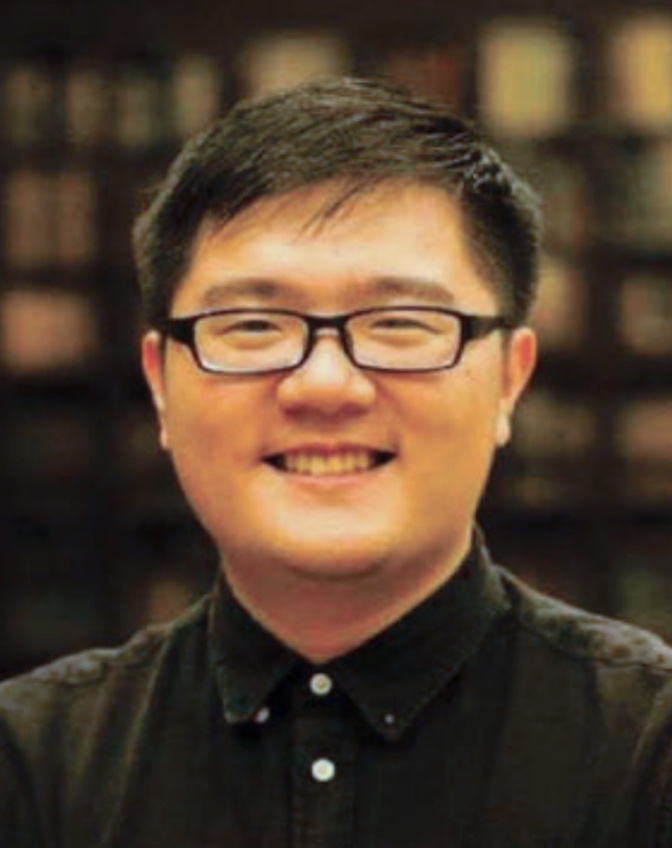}}]{Guo-Wei Yang}
is currently a Ph.D. student in the Department of Computer Science and Technology, Tsinghua University. His research interests include computer graphics, image analysis, and computer vision.
\end{IEEEbiography}

\begin{IEEEbiography}[{\includegraphics[width=1in,height=1.25in,clip,keepaspectratio]{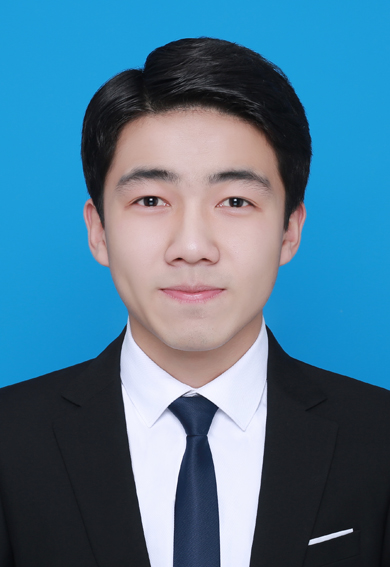}}]{Wen-Yang Zhou}
is currently a Ph.D. student in the Department of Computer Science and Technology, Tsinghua University, Beijing. His research interests include computer graphics, image analysis, and computer vision.
\end{IEEEbiography}

\begin{IEEEbiography}[{\includegraphics[width=1in,height=1.25in,clip,keepaspectratio]{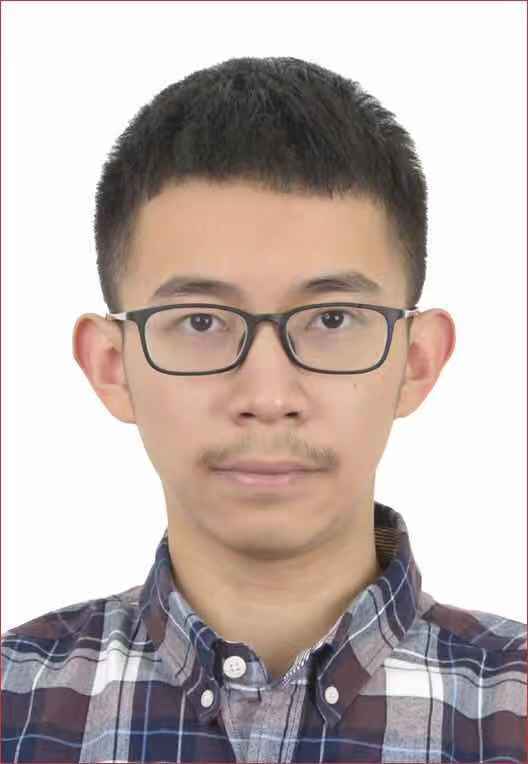}}]{Hao-Yang Peng}
is an undergraduate student at Tsinghua University. His research interests include computer graphics and computer vision.

\end{IEEEbiography}

\begin{IEEEbiography}[{\includegraphics[width=1in,height=1.25in,clip,keepaspectratio]{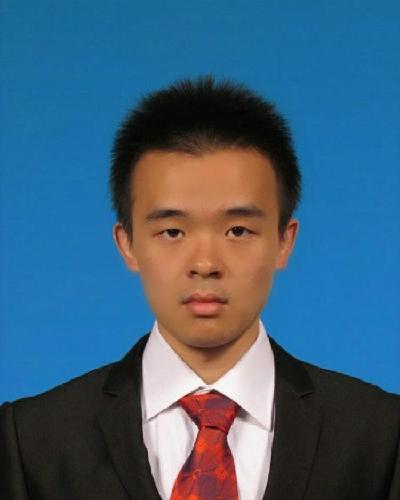}}]{Dun Liang}
is a Ph.D. candidate in the Department of Computer Science and Technology at Tsinghua University, where he received his B.S. degree in computer science and technology, 
in 2016. His research interests include computer graphics, visual media learning and high-performance computing.
\end{IEEEbiography}

\begin{IEEEbiography}[{\includegraphics[width=1in,height=1.25in,clip,keepaspectratio]{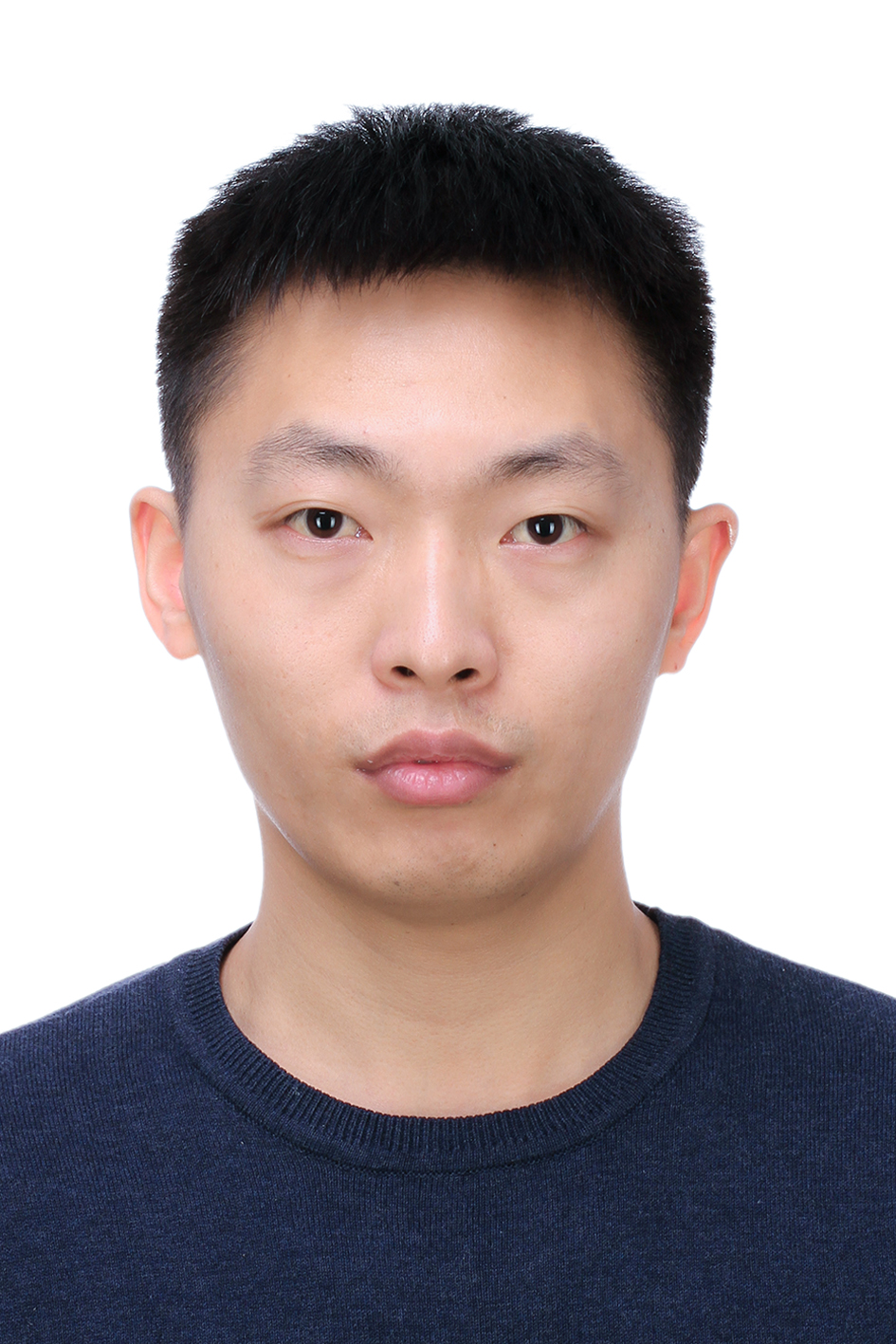}}]{Tai-Jiang Mu}
is currently an assistant researcher at Tsinghua University, where he received his B.S. and Ph.D. degrees in computer science and technology in 2011 and 2016, respectively. His research interests include computer vision, robotics and computer graphics.
\end{IEEEbiography}

\begin{IEEEbiography}[{\includegraphics[width=1in,height=1.25in,clip,keepaspectratio]{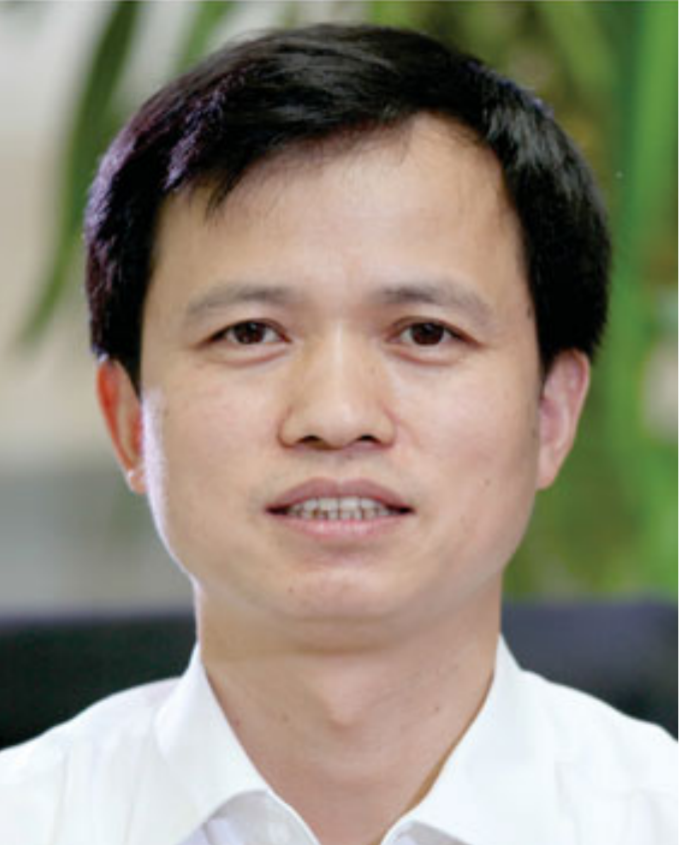}}]{Shi-Min Hu}
received his Ph.D. degree from Zhejiang University, in 1996. He is currently a professor with the Department of Computer Science and Technology, Tsinghua University. He has authored over 100 papers. His research interests include digital geometry processing, video processing, rendering, computer animation, and computer-aided geometric design. He is the Editor-in-Chief of Computational Visual Media, and on the Editorial Board of several other journals, including Computer Aided Design and Computer \& Graphics (both Elsevier). 
\end{IEEEbiography}

\end{document}